\documentclass{article}

\usepackage{microtype}
\usepackage{graphicx}
\usepackage{subfigure}
\usepackage{booktabs} %
\usepackage{amsmath}
\usepackage{tikz}
\usepackage{xspace}
\usepackage{soul}
\usepackage{amssymb}
\usepackage{dblfloatfix}  
\usepackage{enumitem}
\usepackage{booktabs,multirow}

\newcommand{\numcircle}[1]{%
  \tikz[baseline=(char.base)]{
    \node[shape=circle,fill=black,text=white,inner sep=1pt] (char) {\small #1};}}

\usepackage{hyperref}

\usepackage[accepted]{mlsys2025}

\makeatletter
\def\Notice@String{Preprint.}   %
\makeatother

\newcommand{\sys}{\textsc{SinkState}\xspace}
\newcommand{\submissionTitle}{\sys: On-the-Fly SVD of Recurrent States for Efficient Hybrid LLM Serving}

\mlsystitlerunning{\submissionTitle}

\begin{document}

\twocolumn[
\mlsystitle{\submissionTitle}

\mlsyssetsymbol{equal}{*}

\begin{mlsysauthorlist}
\mlsysauthor{Haolin Jiang}{equal,ru}
\mlsysauthor{Weihang Li}{equal,ru}
\mlsysauthor{Zhongzhu Zhou}{usyd}
\mlsysauthor{Jinwen Pan}{ru}
\mlsysauthor{Santosh Pandey}{usf}
\mlsysauthor{Hang Liu}{ru}
\end{mlsysauthorlist}

\mlsysaffiliation{ru}{Rutgers University, New Brunswick, New Jersey, USA}
\mlsysaffiliation{usyd}{University of Sydney, Sydney, New South Wales, Australia}
\mlsysaffiliation{usf}{University of South Florida, Tampa, Florida, USA}

\mlsyscorrespondingauthor{Haolin Jiang}{haolin.jiang@rutgers.edu}
\mlsyscorrespondingauthor{Weihang Li}{wl730@scarletmail.rutgers.edu}

\mlsyskeywords{Linear Attention, Hybrid Model, Low-Rank Compression}

\vskip 0.3in

\begin{abstract}
Hybrid models combine softmax attention with linear attention, which compresses history into fixed-size recurrent states to balance model quality and inference efficiency.
However, linear attention has not pushed this compression to its limit: we find that the queries, keys, and values of linear layers concentrate in a stable, low-dimensional subspace, a phenomenon we call the \emph{linear attention sink}, so each state is effectively low-rank, with a rank that varies across states.
Yet existing serving systems keep these redundant states dense, reading and rewriting them in full at every decoding step, so their memory traffic grows with the batch size and dominates the decoding cost.
Thus \sys stores each recurrent state in an adaptive low-rank form throughout inference with buffered updates and a local SVD restricted to the top subspace.
\sys further consolidates the remaining components asynchronously on the CPU.
On four hybrid models with Mamba-2 and GDN layers, from 4B to 120B parameters, \sys reduces the peak memory of recurrent states by up to 2.6$\times$ (1.9$\times$ on average) and speeds up decoding by up to 1.58$\times$ (1.25$\times$ on average) over Mamba-SSM and FLA, while maintaining accuracy.
\end{abstract}
]

\printAffiliationsAndNotice{\mlsysEqualContribution} %

\section{Introduction}

Linear attention has emerged as a promising solution by compressing the entire history into a fixed-size recurrent state, striking a balance between model quality and inference
efficiency to benefit long sequence applications ranging from reasoning traces~\cite{deepseek2025r1} reaching tens of thousands of tokens to agentic workflows~\cite{yao2023react} spanning hundreds of tool calls~\cite{yang2024sweagent}.
Hybrid models~\cite{lieber2025jamba,minimax2025minimax01,nvidia2025nemotronh,kimi2025linear},
which combine linear and softmax attention layers, have thus gained prominence.
Leading hybrid models like Kimi K3~\cite{kimi2026k3}, Qwen3.8~\cite{qwen2026qwen38}, Nemotron 3 Ultra~\cite{nvidia2025nemotron3},
and GLM-5.3-Flash~\cite{zai2026glm53flash} match full attention baseline at lower inference costs.

However, has linear attention pushed compression to its limit?
The answer is no: across different linear-attention variants, we find that queries, keys, and values concentrate in a low-dimensional subspace that remains stable as the context grows, even for randomly sampled inputs--a phenomenon we call the \emph{linear attention sink} (\S~\ref{sec:motivation:linear_attention_sink}).
This concentration explains the low effective rank observed in recent studies~\cite{nazari2026statereductionlinearattention,sun2026staterankdynamicslinear, menezes2026ghostunmaskingphantomstates}.
Consequently, storing and repeatedly accessing these redundant states in dense form during the decoding wastes GPU memory and bandwidth.
The large state footprint further amplifies this waste: in Qwen3.8-2.4T-A95B under typical settings, recurrent states alone occupy $6{,}144\times$ the KV-cache memory per token.

Existing approaches do not fully exploit the redundancy within each state. 
They reduce this overhead through state reuse~\cite{huang2026unifiedradix,pan2025marconi,liu2026suffixreplay} or deferred state writes ~\cite{liou2026replayssm, pan2026leapquant}, but all retain dense representations.
State pruning~\cite{nazari2026statereductionlinearattention,menezes2026ghostunmaskingphantomstates,sun2026staterankdynamicslinear} and calibrated projections~\cite{kwon2026sketchssm} largely rely on static compression choices, with no guarantee of generalization to unseen inputs.

A straightforward way to exploit this redundancy is to recompute the SVD of each state after every update.
However, this is prohibitively expensive: at a batch size of 128, running SVD on all states of Nemotron-3-Nano-4B takes 4,000~ms, over two orders of magnitude longer than a typical decoding step (20--30~ms).
We therefore ask: \emph{can we exploit this redundancy on the fly by adaptively discarding unimportant state directions, reducing memory usage and decoding latency while preserving model quality?}

Building upon the linear attention sink, we make two key observations.
\ul{Observation i:} Incoming keys and values are mainly captured by the leading components of the state, with diminishing contributions toward the tail.
This suggests that the factorization can be maintained through inexpensive local updates.
\ul{Observation ii:} When the remaining components are left unchanged over many updates, similar directions may be stored as separate components, inflating the stored rank and memory footprint.
Periodic consolidation is therefore needed for memory and bandwidth savings.

We present \sys, a serving system that adaptively compresses recurrent states throughout inference using three techniques.
\ul{i. Local SVD Update} (\S\ref{sec:design:local_svd_update}).
For a given state, \sys splits its existing factors into an \emph{active} part containing the leading components and a \emph{frozen} part containing the rest.
It computes a small local SVD on the active part together with new token updates, while keeping the frozen part unchanged.
To further amortize the cost, \sys buffers $K$ token updates and merges them at once.
\ul{ii. Asynchronous CPU Consolidation} (\S\ref{sec:design:async_consolidation}).
\sys asynchronously consolidates the frozen factors via an exact SVD on the CPU, removing redundant components while GPU decoding proceeds.
As CPU consolidation cannot keep pace with GPU compactions, \sys processes layers in a round-robin order.
\ul{iii. Kernel Design} (\S\ref{sec:design:kernel}).
\sys uses a compact memory layout to accommodate ranks that differ across states and change during decoding.
Fused kernels append incoming updates to the buffer and compute outputs directly from the factors and buffered updates, without materializing dense states.
\sys also optimizes load balancing and supports CUDA Graphs.

To summarize, we make the following three contributions:
\begin{itemize}
\item We identify the \emph{linear attention sink} across Mamba-2, GDN, and KDA, which explains the low effective rank of recurrent states (\S\ref{sec:motivation:linear_attention_sink}).
\item We design \sys, which tracks the dominant subspace of each state online with local SVD and asynchronous CPU consolidation, supported by optimized GPU kernels (\S\ref{sec:design}).
\item We conduct extensive experiments on four hybrid models, showing
that \sys reduces the peak recurrent-state memory by
1.6--2.6$\times$, improves end-to-end decoding throughput by up to
1.58$\times$, accelerates the state kernels by up to 2.41$\times$, and
serves up to 2$\times$ larger batches, while maintaining model
accuracy (\S\ref{sec:eval}).

\end{itemize}

\section{Background}
\label{sec:background}

\subsection{Linear Attention and Hybrid Models}
\label{sec:bg-hybrid}

Linear attention, or the state space model (SSM), replaces softmax attention with a recurrence over a fixed-size matrix, called the state $H_t$~\cite{katharopoulos2020transformers,gu2024mamba,dao2024mamba2}.
At the $t$-th token, linear attention produces three vectors, a query $q_t$, a key $k_t$, and a value $v_t$.
The key $k_t$ and value $v_t$ update the previous state $H_{t-1}$ into the current state $H_t$, and the query $q_t$ then computes the output $y_t$, also called the readout, from $H_t$:
\begin{equation}
    H_t=\alpha_tH_{t-1}+v_tk_t^\top,
    \label{eq:update}
\end{equation}
\begin{equation}
    y_t=H_tq_t,
    \label{eq:output}
\end{equation}
where $\alpha_t\in(0,1)$ is a scalar decay, as in Mamba2~\cite{dao2024mamba2}.
Later variants change only how the state forgets, i.e., the first term of Equation~\eqref{eq:update}, and keep the rank-1 write $v_tk_t^\top$.
GDN~\cite{yang2025gdn} applies the delta rule~\cite{yang2024deltanet}, which replaces $\alpha_tH_{t-1}$ with $\alpha_tH_{t-1}(I-\beta_tk_tk_t^\top)$ and scales the write by $\beta_t\in(0,1)$.
KDA~\cite{kimi2025linear} further replaces the scalar $\alpha_t$ with a diagonal matrix $\mathrm{Diag}(\alpha_t)$ applied to the columns of $H_{t-1}$, giving $H_{t-1}\mathrm{Diag}(\alpha_t)(I-\beta_tk_tk_t^\top)$.
Overall, all of these variants follow the same process: (i) a decay of $H_{t-1}$ plus a rank-1 update $v_tk_t^\top$, and (ii) the output $y_t=H_tq_t$.
Hybrid models stack many such layers with a few full-attention layers~\cite{lieber2025jamba,minimax2025minimax01,nvidia2025nemotronh,qwen2026qwen35,kimi2025linear}.

\begin{figure*}[!t]
    \centering
    \begin{minipage}[t]{0.49\linewidth}
        \centering
        \includegraphics[width=\linewidth]{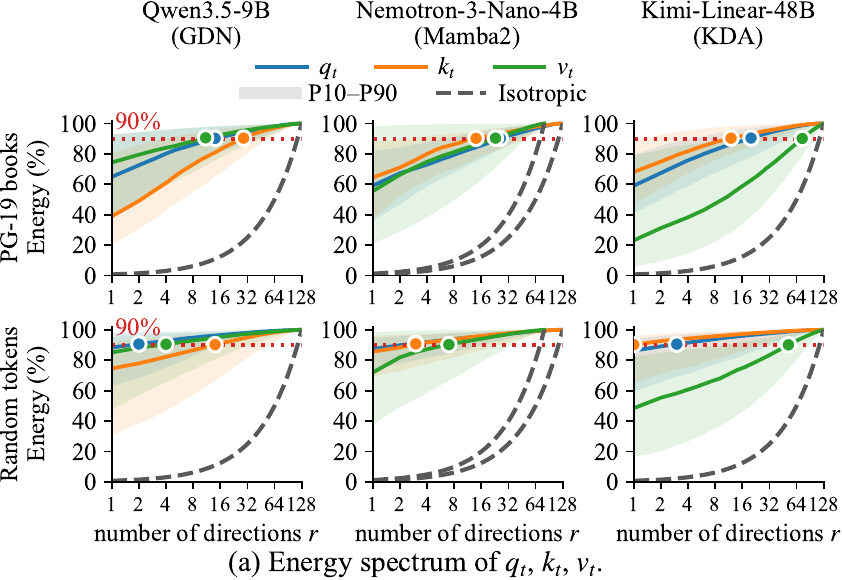}
    \end{minipage}\hfill
    \begin{minipage}[t]{0.49\linewidth}
        \centering
        \includegraphics[width=\linewidth]{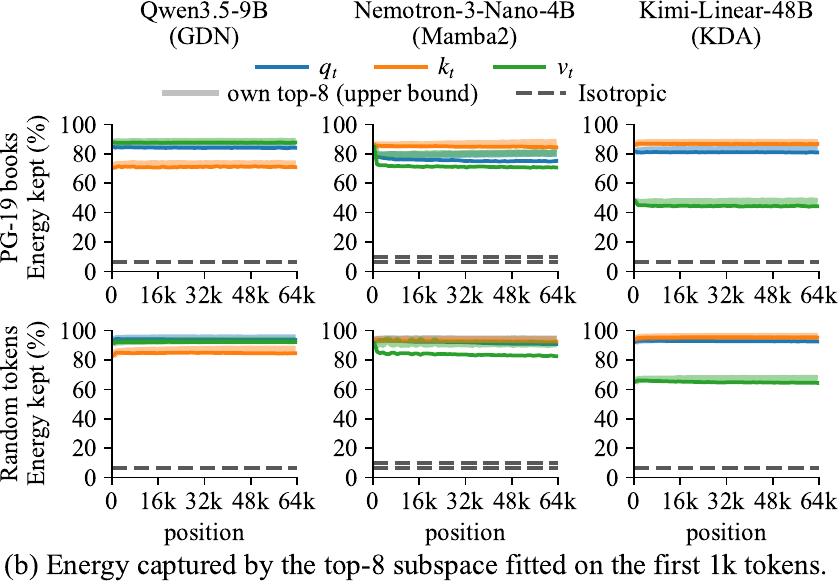}
    \end{minipage}

    \vspace{4pt}

    \caption{The $q_t$, $k_t$, and $v_t$ vectors of three linear-attention models on PG-19 (top rows) and uniformly random tokens (bottom rows), profiled on 32 sequences of 64k tokens for each state $H_t$. 
    (a) Energy spectrum: the fraction of energy captured by the top-$r$ directions of the stacked $q_t$, $k_t$, and $v_t$.
    (b) Stability: energy of each segment $[n\text{k},(n+1)\text{k})$ captured by the top-8 subspace computed on $[0,1\text{k})$, versus its own optimal top-8 subspace.
    Gray dashed lines show the results for uniformly random (isotropic) directions.}
    \label{fig:qkv_motivation}
\end{figure*}

\subsection{Low-Rank Factorization}
\label{sec:bg-svd}

The singular value decomposition (SVD) writes a matrix as
$H=U\Sigma V^\top$ with singular values $\sigma_0\ge\sigma_1\ge\cdots$,
and truncating the small ones gives the best low-rank approximation of
$H$ in the Frobenius norm.
Since an exact SVD of a $d\times d$ matrix costs $O(d^3)$, randomized
SVD~\cite{halko2010findingstructurerandomnessprobabilistic,martinsson2015randomizedblockedalgorithmefficiently}
approximates the dominant subspace with a few random projections, and
its blocked variants determine the rank on the fly.
Incremental SVD~\cite{brand2006fast} further updates a factorization
after a low-rank modification by decomposing a small core matrix.
In LLM inference, SVD compresses static weights~\citep{wang2025svdllm} and KV caches~\citep{chang2026xkv} once before serving.
A recurrent state, however, changes at every decoding step, and to our
knowledge, no prior work keeps it in a low-rank SVD form throughout
decoding.

\section{Motivation}
\label{sec:motivation}

\subsection{Linear Attention Sink}
\label{sec:motivation:linear_attention_sink}

We find that, across different linear-attention variants, the queries $q_t$, keys $k_t$, and values $v_t$ concentrate in a stable and semantics-independent low-dimensional subspace, making both the state $H_t$ and its output $H_tq_t$ dominated by a few directions, a phenomenon we call the \emph{linear attention sink}.

Figure~\ref{fig:qkv_motivation} profiles Qwen3.5-9B (GDN), Nemotron-3-Nano-4B (Mamba2), and Kimi-Linear-48B (KDA) on 32 sequences of 64k tokens (1k = 1024) from PG-19 and from random tokens.
For each state, we stack all its $q_t$, $k_t$, and $v_t$ vectors into three matrices.
Figure~\ref{fig:qkv_motivation}(a) measures the fraction of each matrix captured by its top-$r$ subspace, and Figure~\ref{fig:qkv_motivation}(b) profiles whether this top-$r$ subspace is stable as the context grows to 64k tokens.
On PG-19 (Figure~\ref{fig:qkv_motivation}(a)), the top-1 direction captures 39--74\% of each stacked matrix ($44$--$95\times$ more than isotropic vectors) and the top-8 directions capture 73--88\% ($8$--$14\times$ more) in most cases.
Figure~\ref{fig:qkv_motivation}(b) shows that this dominant subspace remains stable as the context grows: it still accounts for 70.8--87.8\% even at 64k, reaching 87.6--99.2\% of an optimal subspace recomputed there.
More importantly, neither the concentration nor its stability relies on the semantics of the given context.
On random tokens, this phenomenon becomes even stronger: the top-1 and top-8 directions capture 72--90\% and 87--96\% (Figure~\ref{fig:qkv_motivation}(a)), and the top-8 directions still keep 82.6--95.3\% at 64k (90.8--99.2\% of the optimal, Figure~\ref{fig:qkv_motivation}(b)).

\begin{figure}[htbp]
    \centering
    \includegraphics[width=\linewidth]{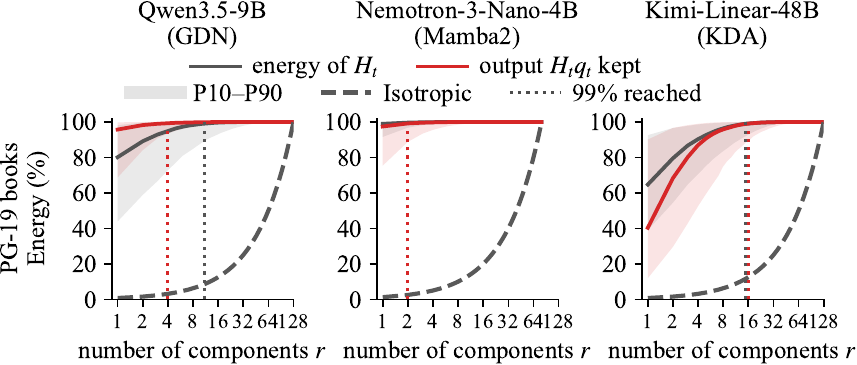}
    \vspace{-16pt}
    \caption{Fractions of $H_t$'s energy and output $H_tq_t$ retained when only the top-$r$ components of $H_t$ are kept. Measurements are taken every 0.5k tokens across 32 sequences of 64k tokens each.}
    \label{fig:state_readout}
\end{figure}

Since $H_t$ accumulates $v_tk_t^\top$ at each step, the dominant subspace of $k_t$ and $v_t$ piles up into a few top components of $H_t$, which also contribute most of the output $H_tq_t$.
As shown in Figure~\ref{fig:state_readout}, only the top 2--15 of up to 128 components already contribute 99\% of $H_t$, and keeping only the top 2--16 components recovers 99\% of the output $H_tq_t$.
Therefore, the dense state $H_t$ is highly redundant: a few top components already recover almost all of $H_t$ and its output, yet every decoding step still reads and writes the full dense state, wasting both GPU memory and bandwidth and increasing the latency of each step.
A straightforward approach is to project $H_t$ onto an offline-calibrated low-rank basis, which is neither general nor scalable.
This factorization therefore has to be maintained online, yet recomputing an exact SVD at every decoding step is prohibitive.
For example, at a batch size of 128, Nemotron-3-Nano-4B holds 258k recurrent states of size $80\times128$ (21 Mamba2 layers $\times$ 96 heads $\times$ 128 requests), and running an exact SVD on all of them takes about 4000~ms, over $100\times$ longer than a typical decoding step ($\sim$20--30~ms).
Thus, \sys builds upon two key observations to efficiently keep each state factorized during decoding (Section~\ref{sec:motivation:on_the_fly_svd}).

\subsection{Enabling On-the-Fly SVD}
\label{sec:motivation:on_the_fly_svd}

\begin{figure}[htbp]
    \centering
    \includegraphics[width=\linewidth]{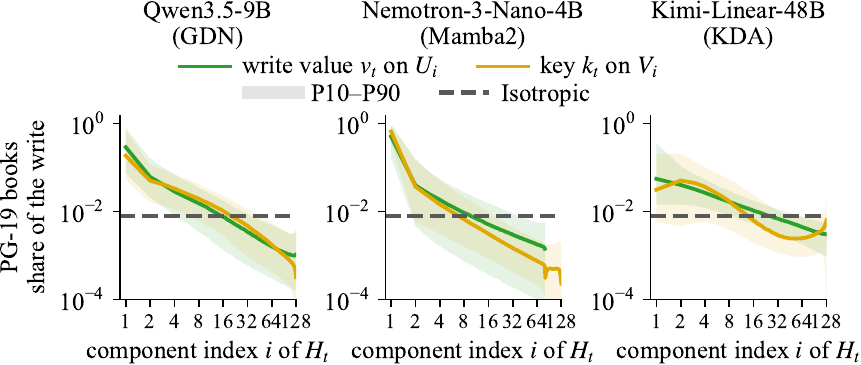}
    \caption{Share of an incoming $v_t$, $k_t$ on each component of $H_{t-1}$.}
    \label{fig:write_share}
\end{figure}

\textbf{Observation 1: The incoming $k_t$ and $v_t$ are mostly captured by the top components of $H_{t-1}$, with a diminishing share toward the tail.}
Given the factorized form $H_{t-1}=U_{t-1}\Sigma_{t-1}V_{t-1}^\top$ obtained at step $t-1$, Figure~\ref{fig:write_share} measures the share of each incoming $v_t$ captured by the $i$-th column of $U_{t-1}$, i.e., $(U_{t-1}[i]^\top v_t)^2/\|v_t\|^2$, and likewise for $k_t$ and $V_{t-1}$.
On Qwen3.5 and Nemotron-3-Nano, the top-1 column already captures 18--65\% of $v_t$ and $k_t$ ($23$--$83\times$ more than isotropic vectors, which capture $\sim$1\% per column), and the top-8 columns capture 40--76\% ($6.3$--$12\times$ more).
Kimi-Linear follows the same trend, with 29--40\% in its top-8 columns ($4.6$--$6.4\times$ more).
In contrast, the later columns (e.g., $>$16 on Qwen3.5) capture even less than isotropic vectors, with the share of the last column diminishing to 0.03--0.1\%.

\textbf{Design 1: Merging new tokens into only the top subspace keeps SVD practical on the fly.}
Given the factorized form $H_{t-1}=U_{t-1}\Sigma_{t-1}V_{t-1}^\top$, \sys splits it into an \emph{active} part (the top-$r$ columns $U_{t-1}[:r]$ and $V_{t-1}[:r]$), and a \emph{frozen} part (the remaining $U_{t-1}[r:]$ and $V_{t-1}[r:]$).
It concatenates only the new tokens with the active part into a small matrix and runs a \emph{local SVD} on it, giving the updated singular vectors $U_{\mathrm{active}}, V_{\mathrm{active}}$ and singular values $\sigma_0\ge\sigma_1\ge\cdots$.
It then merges $U_{\mathrm{active}}, V_{\mathrm{active}}$ back with the frozen part $U_{t-1}[r:], V_{t-1}[r:]$, and drops the components below $\varepsilon\sigma_0$, where $\varepsilon$ is a threshold.
To further amortize the cost of local SVD, \sys buffers $K$ new tokens and merges them at once (Section~\ref{sec:design:local_svd_update}).

\begin{figure}[htbp]
    \centering
    \includegraphics[width=\linewidth]{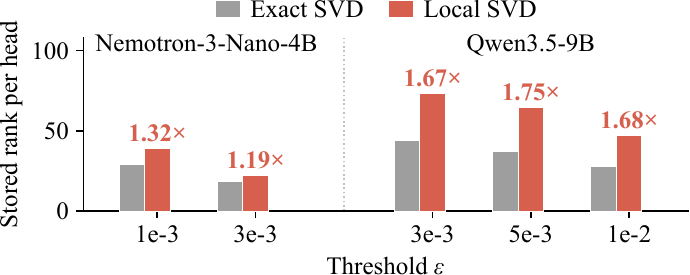}
    \vspace{-16pt}
    \caption{Rank stored per head under local SVD and exact SVD after 128 decoding steps on GSM8K.}
    \label{fig:frozen_inflation}
\end{figure}

\textbf{Observation 2: With local SVD alone, redundant components accumulate in the frozen part and inflate the memory footprint.}
Since the frozen part never enters the local SVD, it may continue to accumulate components with similar directions, making the stored rank much larger than that of an exact SVD.
As shown in Figure~\ref{fig:frozen_inflation}, local SVD stores $1.19$--$1.32\times$ more components per head than an exact SVD on Nemotron-3-Nano and $1.67$--$1.75\times$ more on Qwen3.5 across thresholds, offsetting the savings from the low-rank factorization.
However, simply letting the local SVD process more columns slows down decoding.

\textbf{Design 2: Consolidating the frozen part asynchronously on idle CPUs keeps it off the critical path.}
During decoding, the GPU runs the model while the host CPUs stay mostly idle ($\sim$2\% utilization).
\sys consolidates the frozen part asynchronously on the CPU: since it stays unchanged between two consecutive SVD compactions, \sys downloads it, merges its redundant components with an OpenMP-parallel exact SVD, and uploads the result back before the next compaction. 
As the CPU cannot keep pace with the GPU compactions, \sys consolidates the layers in a round-robin order (Section~\ref{sec:design:async_consolidation}).

\section{Design}
\label{sec:design}

\subsection{Overview}
\label{sec:design:overview}

\begin{figure}[htbp]
    \centering
    \includegraphics[width=\linewidth]{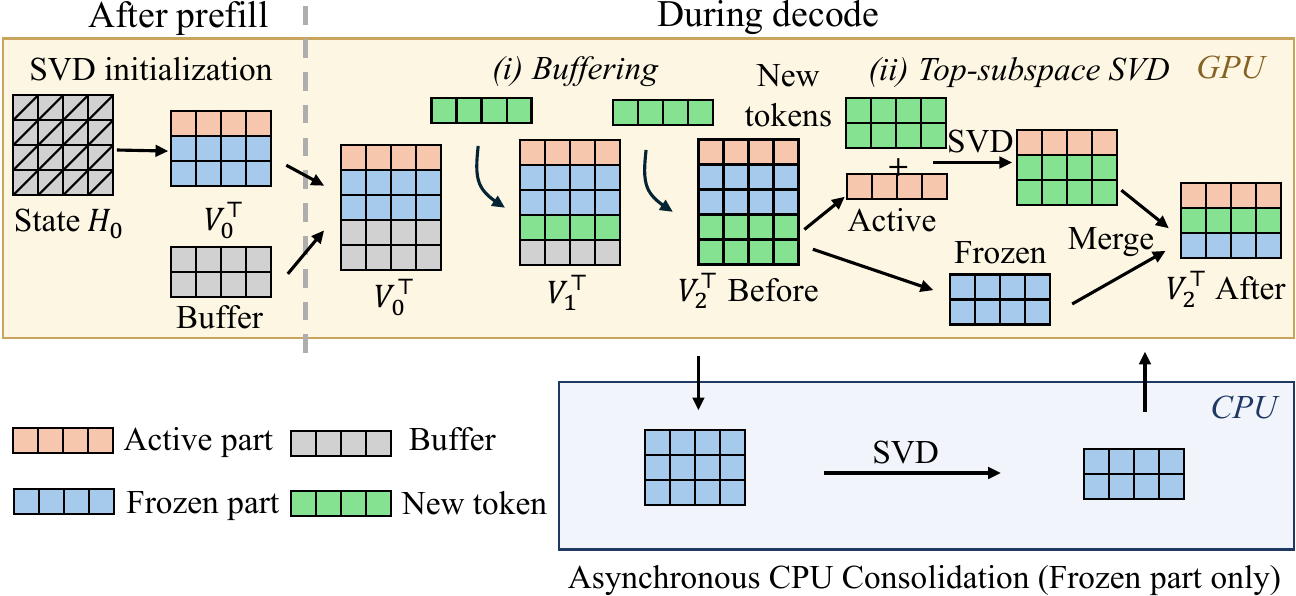}
    \vspace{-16pt}
    \caption{Overview of \sys.}
    \label{fig:overview}
\end{figure}

\begin{figure*}[htbp]
    \centering
    \includegraphics[width=\linewidth]{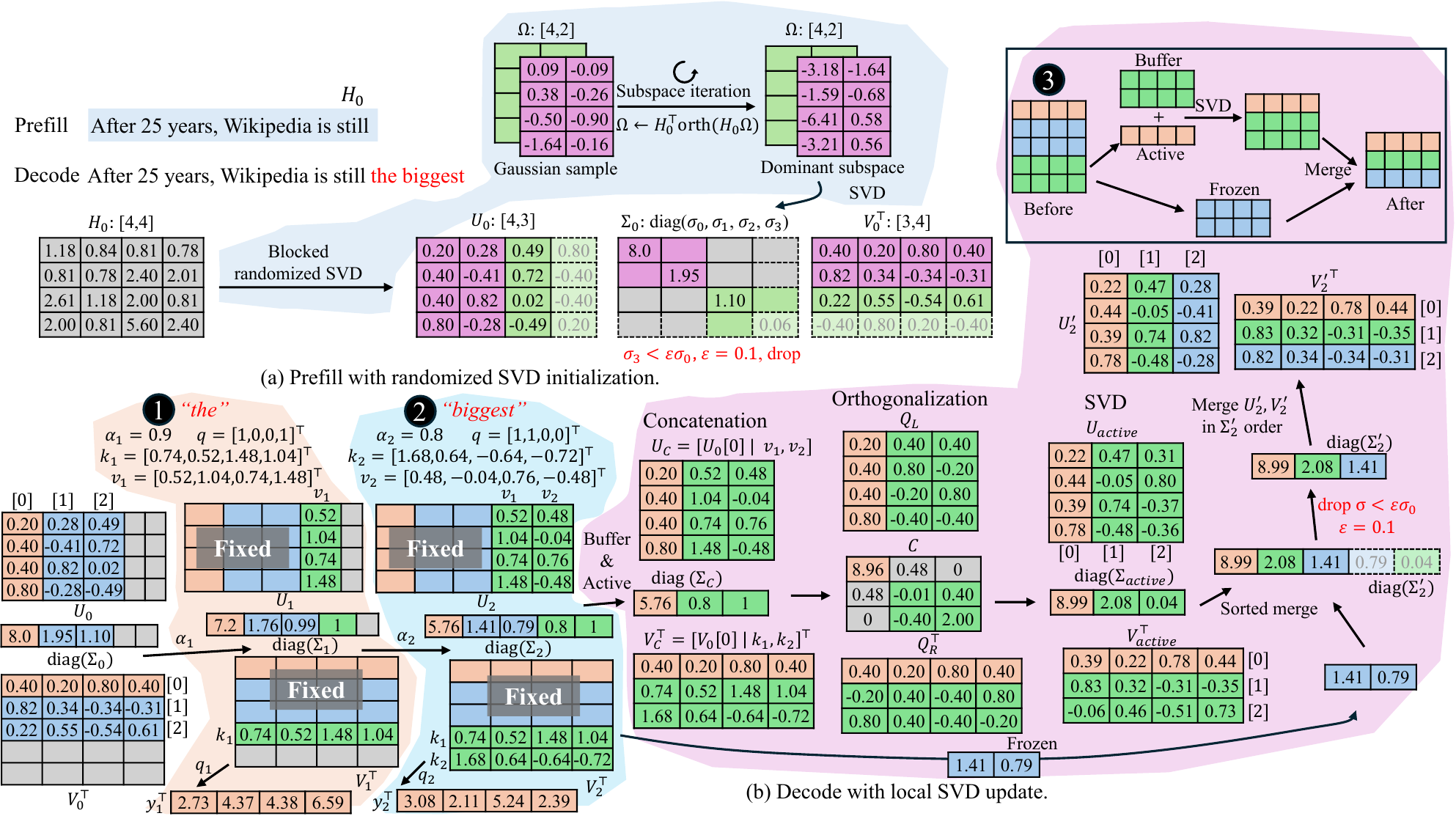}
    \vspace{-16pt}
    \caption{Prefill and decoding with local SVD updates.}
    \label{fig:local_svd_update}
\end{figure*}

The goal of \sys is to adaptively maintain the rank of each state, so that states that carry more information retain a higher rank, while the others retain a lower rank.
To determine which components to retain, \sys drops those whose singular values ($\sigma$) fall below a fraction $\varepsilon$ of the largest singular value ($\sigma_0$), i.e., $\sigma<\varepsilon\sigma_0$.
\sys applies this truncation at initialization and at every compaction.
\textbf{After prefill}, \sys first initializes the low-rank factors of the resulting state, $H_0\approx U_0\Sigma_0V_0^\top$.
To avoid the prohibitive cost of an exact SVD, \sys adopts the widely used randomized SVD~\cite{halko2010findingstructurerandomnessprobabilistic, martinsson2015randomizedblockedalgorithmefficiently} and computes it block by block.
\textbf{During decode}, \sys maintains the low-rank factors $U_0,V_0$ in two ways:
\ul{(i) \emph{Buffering}}: to amortize the cost of SVD across decoding steps, \sys caches the keys and values of new tokens in a buffer of size $K$ and compacts the factors only when the buffer is full.
\ul{(ii) \emph{Top-subspace SVD}}: as new updates are mainly absorbed by the top subspace (see Section~\ref{sec:motivation:on_the_fly_svd}), \sys merges the buffer only into this subspace, while the other components remain frozen, thereby avoiding the prohibitive cost of running an SVD on the whole state.

\subsection{Local SVD Update}
\label{sec:design:local_svd_update}

Figure~\ref{fig:local_svd_update} illustrates the local SVD update on a single $4\times4$ state $H_0$, using $\varepsilon=0.1$ and the prompt \textit{``After 25 years, Wikipedia is still''} followed by two generated tokens, \textit{``the''} and \textit{``biggest''}. This example uses buffer size $K=2$.

\textbf{Randomized SVD initialization after prefill.}
After prefilling the prompt ``After 25 years, Wikipedia is still'', \sys obtains the $4\times4$ state $H_0$ (gray matrix in Figure~\ref{fig:local_svd_update}(a)), whose first row is $[1.18,0.84,0.81,0.78]$, and approximates it as $H_0\approx U_0\Sigma_0V_0^\top$ using randomized SVD.
Specifically, \sys constructs the approximation one block at a time through \ul{(i) subspace approximation} and \ul{(ii) block SVD}, and checks the resulting singular values against the threshold to decide whether to continue.

\ul{(i) Subspace approximation} identifies the dominant subspace of the state $H_0$ using a small random matrix.
Specifically, \sys first samples a $4\times2$ Gaussian matrix $\Omega$ (purple matrix in Figure~\ref{fig:local_svd_update}(a)) with $[0.09,-0.09]$ as its first row.
The matrix $\Omega$ is then refined through repeated subspace iteration~\cite{halko2010findingstructurerandomnessprobabilistic}:
$\Omega\leftarrow H_0^\top\mathrm{orth}(H_0\Omega)$.
Each iteration multiplies $\Omega$ by $H_0$, orthogonalizes the result, and multiplies it by $H_0^\top$.
After a few iterations, $\Omega$ approximately spans the top-$2$ subspace of $H_0$, giving $[-3.18,-1.64]$ as its first row.

\ul{(ii) Block SVD} then extracts the singular components of the current block from $\Omega$.
Specifically, \sys applies SVD to $\Omega$, yielding the singular values $8.0$ and $1.95$ and their corresponding singular vectors, shown as the pink columns of $U_0$ (first row $[0.20,0.28]$) and pink rows of $V_0^\top$ (first row $[0.40,0.20,0.80,0.40]$).

Since both singular values exceed the threshold $\varepsilon\sigma_0=0.1\times8.0=0.8$, $H_0$ may still contain significant directions, so \sys repeats the two steps on a new block.
It samples another random matrix, finds the next dominant subspace (green matrix), and applies SVD, yielding singular values $1.10$ and $0.06$ along with their singular vectors (the green columns of $U_0$ and the green rows of $V_0^\top$).
Since $0.06$ falls below $0.8$, \sys truncates it (dashed) and stops adding blocks.
Finally, \sys concatenates the kept components of both blocks into $U_0$ (first row $[0.20,0.28,0.49]$), $\Sigma_0=\mathrm{diag}(8.0,1.95,1.10)$, and $V_0$ (first row $[0.40,0.20,0.80,0.40]$), giving a rank-$3$ approximation $H_0\approx U_0\Sigma_0V_0^\top$.

\textbf{Decode with local SVD update.}
\sys then decodes two tokens, \textit{``the''} and \textit{``biggest''} (Figure~\ref{fig:local_svd_update}(b)).
For ease of illustration, the figure uses the Mamba2 recurrence~\cite{dao2024mamba2}.
The $t$-th token produces a key $k_t$, a value $v_t$, a query $q_t$, and a scalar decay $\alpha_t\in(0,1)$.
The state is then updated as $H_t=\alpha_tH_{t-1}+v_tk_t^\top$, and the output is $y_t=H_tq_t$.
The same procedure also applies to GDN~\cite{yang2025gdn}.
\sys maintains the low-rank factors through \ul{(i) buffering} and \ul{(ii) top-subspace SVD}.

\ul{(i) Buffering} caches the new $v_t$ and $k_t$ of each token and applies the decay $\alpha_t$ to the singular values $\Sigma_0$, keeping $U_0$ and $V_0$ unchanged (marked as fixed in Figure~\ref{fig:local_svd_update}(b)).
At \numcircle{1} (orange region), \sys processes the token \textit{``the''} with $\alpha_1=0.9$, $q_1=[1,0,0,1]^\top$, $k_1=[0.74,0.52,1.48,1.04]^\top$, and $v_1=[0.52,1.04,0.74,1.48]^\top$.
\sys first applies the decay $\alpha_1$ to the singular values $\Sigma_0$, scaling $\mathrm{diag}(8.0,1.95,1.10)$ to $\mathrm{diag}(7.2,1.76,0.99)$.
It then appends $v_1$ as a new column of $U_0$ and $k_1$ as a new row of $V_0^\top$, filling the first buffer slot (gray slots turn green) and giving $U_1=[U_0\mid v_1]$ and $V_1^\top=[V_0\mid k_1]^\top$.
To match the extended factors, \sys also appends a weight of $1$ to the singular values, giving $\mathrm{diag}(\Sigma_1)=[7.2,1.76,0.99,1]$ and completing the factors $U_1\Sigma_1V_1^\top$.
The output $y_1$ is then computed from the extended factors with the query $q_1=[1,0,0,1]^\top$, giving $y_1=U_1\Sigma_1V_1^\top q_1=[2.73,4.37,4.38,6.59]^\top$.
At \numcircle{2} (blue region), \sys processes the token \textit{``biggest''} with $\alpha_2=0.8$, $q_2=[1,1,0,0]^\top$, $k_2=[1.68,0.64,-0.64,-0.72]^\top$, and $v_2=[0.48,-0.04,0.76,-0.48]^\top$.
In the same way, \sys first applies the decay $\alpha_2$ to all singular values, including the buffered weight, scaling $[7.2,1.76,0.99,1]$ to $[5.76,1.41,0.79,0.8]$.
It then appends $v_2$ and $k_2$ to the second buffer slot, giving $U_2=[U_1\mid v_2]$, $V_2^\top=[V_1\mid k_2]^\top$, and $\mathrm{diag}(\Sigma_2)=[5.76,1.41,0.79,0.8,1]$.
The output is computed as $y_2=U_2\Sigma_2V_2^\top q_2=[3.08,2.11,5.24,2.39]^\top$.
Now the buffer is full, triggering the compaction in \numcircle{3}.

\ul{(ii) Top-subspace SVD} aims to compact the extended factors with a small, local SVD, thereby avoiding the prohibitive overhead of running an SVD over all the factors.
The black box at \numcircle{3} gives a brief sketch of this procedure.
\sys first splits the current factors into an \emph{active} part, i.e., the dominant components with the largest singular values (orange), and a \emph{frozen} part with the remaining components (blue).
It then concatenates the active part with the buffer (green) and runs a small SVD on them.
Finally, \sys merges the frozen part into the result and drops the components below the threshold.
In this example, the active part is the top-1 component ($5.76$) with $U_0[0]$ and $V_0[0]$, and the frozen part keeps $1.41$ and $0.79$.
As shown in the pink region, \sys concatenates the active part with the buffer into $U_C = [U_0[0]\mid v_1,v_2]$, $\Sigma_C = \mathrm{diag}(5.76,0.8,1)$, and $V_C = [V_0[0]\mid k_1,k_2]$.
To keep the SVD small, \sys uses an equivalent reformulation following Brand's incremental SVD~\cite{brand2006fast} to decompose only a small core matrix instead of the concatenated factors.
By orthogonalizing the two concatenated factors into $Q_L$ and $Q_R^\top$, \sys rewrites their product with the diagonal factor as $U_C\Sigma_CV_C^\top=Q_LCQ_R^\top$, where only the small $3\times3$ core $C$ (first row $[8.96,0.48,0]$) requires an SVD.
An SVD of $C$ then yields $U_{\mathrm{active}}$ (first row $[0.22,0.47,0.31]$), $\mathrm{diag}(\Sigma_{\mathrm{active}})=[8.99,2.08,0.04]$, and $V_{\mathrm{active}}^\top$ (first row $[0.39,0.22,0.78,0.44]$).
\sys then merges these singular values with the frozen ones $[1.41, 0.79]$ in descending order, giving $[8.99,2.08,1.41,0.79,0.04]$, and drops those below $\varepsilon\sigma_0=0.1\times8.99=0.899$ (dashed), giving $\mathrm{diag}(\Sigma_2')=[8.99,2.08,1.41]$.
\sys then concatenates the corresponding singular vectors from $U_{\mathrm{active}}$, $V_{\mathrm{active}}$, and the frozen parts (blue slots) in the order of $\mathrm{diag}(\Sigma_2')$ to form $U_2'$ and $V_2'$.

\textbf{Staggered compaction across layers.}
If all layers compact their buffers at the same step, that step runs all the local SVDs at once, causing a latency spike every $K$ steps.
\sys therefore staggers the compaction across layers: at each decoding step, only $1/K$ of the layers compact their buffers, and each layer is offset by a different number of steps.
For example, in a $24$-layer model with $K=8$, three layers are compacted at each step, thereby amortizing the compaction cost evenly across decoding steps.

\subsection{Asynchronous CPU Consolidation}
\label{sec:design:async_consolidation}

\begin{figure}[htbp]
    \centering
    \includegraphics[width=\linewidth]{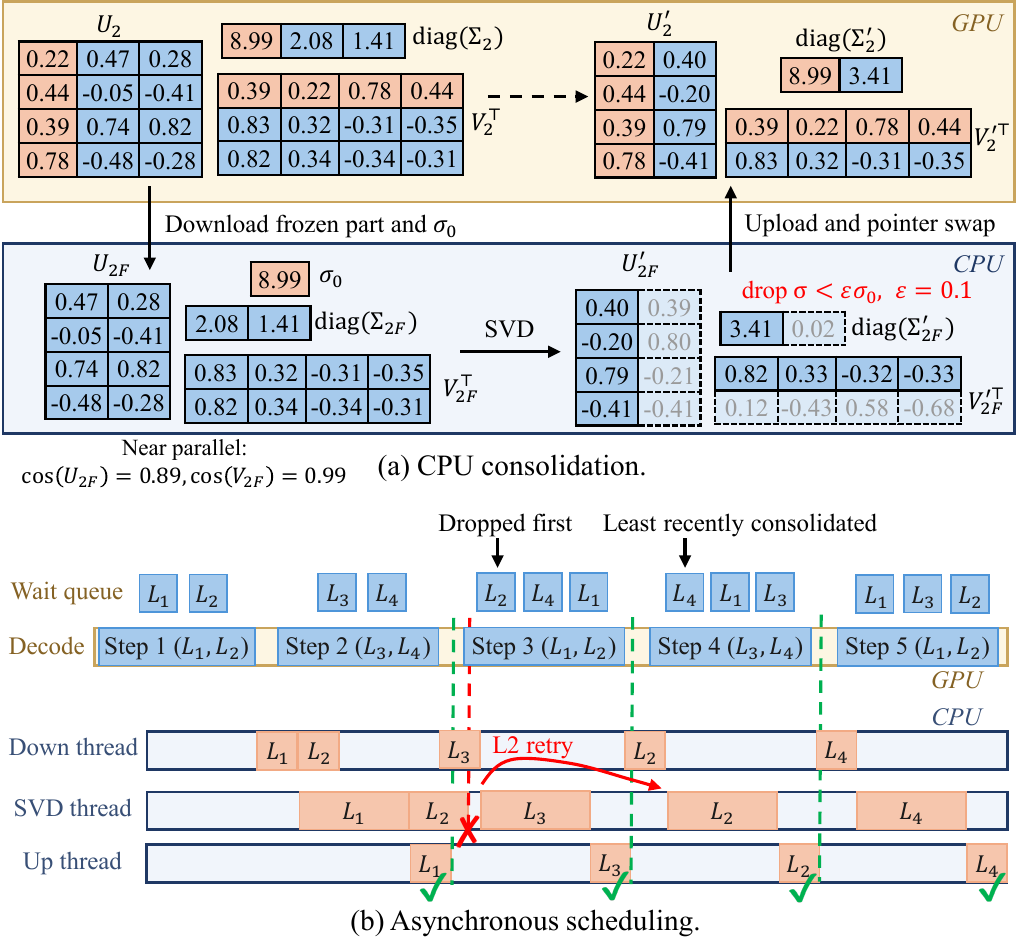}
    \vspace{-16pt}
    \caption{Asynchronous CPU Consolidation.}
    \label{fig:async_consolidation}
\end{figure}

Local SVD keeps each update small by leaving the frozen part untouched. 
However, later updates may re-add existing directions as new frozen rows, causing the number of frozen rows to gradually exceed the actual rank of the frozen part and increasing both GPU memory usage and memory traffic during decoding.
\sys therefore periodically consolidates the frozen part with an exact SVD on the CPU.
Figure~\ref{fig:async_consolidation} illustrates this process in two parts: \textbf{(i) CPU consolidation} of one layer (Figure~\ref{fig:async_consolidation}(a)), and \textbf{(ii) Asynchronous scheduling} of the consolidation across layers (Figure~\ref{fig:async_consolidation}(b)).

\textbf{CPU consolidation} re-decomposes the frozen part to eliminate redundant components.
Figure~\ref{fig:async_consolidation}(a) continues the example of Figure~\ref{fig:local_svd_update}, where the factors after the local SVD update are denoted as $U_2$, $\Sigma_2$, and $V_2$ for brevity.
Here $U_2$ has $[0.22,0.47,0.28]$ as its first row, $V_2^\top$ has $[0.39,0.22,0.78,0.44]$ as its first row, and $\mathrm{diag}(\Sigma_2)=[8.99,2.08,1.41]$, with only the largest component ($8.99$, orange) active.
The two frozen components $U_{2F}, V_{2F}$ with singular values $2.08$ and $1.41$ have similar directions, with $\mathrm{cos}(U_{2F})=0.89$ and $\mathrm{cos}(V_{2F})=0.99$.
\sys first downloads the frozen part $U_{2F}$, $\Sigma_{2F}$, and $V_{2F}$ to the CPU, together with the largest singular value $\sigma_0=8.99$.
An exact SVD is then performed on the frozen part, yielding $U_{2F}'$, $\mathrm{diag}(\Sigma_{2F}')=[3.41,0.02]$, and $V_{2F}'$.
Since $0.02$ falls below $\varepsilon\sigma_0=0.1\times8.99=0.899$, \sys drops it (dashed) and keeps a single component, i.e., the first column of $U_{2F}'$, $[0.40,-0.20,0.79,-0.41]^\top$, the singular value $3.41$, and the first row of $V_{2F}'^\top$, $[0.82,0.33,-0.32,-0.33]$.
Finally, \sys uploads the retained frozen factors to the GPU and replaces the unconsolidated frozen part, yielding $\mathrm{diag}(\Sigma_2')=[8.99,3.41]$ with corresponding factors $U_2'$ and $V_2'^\top$. This reduces the rank from $3$ to $2$ while leaving the active part untouched on the GPU. 
In practice, the replacement is implemented by swapping pointers, avoiding additional data copies on the GPU.
For layers with multiple heads, \sys uses OpenMP to parallelize consolidation across heads.

\textbf{Asynchronous scheduling} overlaps the CPU consolidation with GPU decoding without slowing down decoding.
Figure~\ref{fig:async_consolidation}(b) shows a model with four layers ($L_1$--$L_4$), where staggered compaction (Section~\ref{sec:design:local_svd_update}) compacts two layers at each decoding step, e.g., $L_1$ and $L_2$ at Step~1 and $L_3$ and $L_4$ at Step~2.
After each decoding step, the compacted layers enter a wait queue and are processed by a three-stage CPU pipeline, with dedicated threads for downloading the frozen part, performing SVD, and uploading the result.
In this example, Step~1 compacts $L_1$ and $L_2$, so both enter the wait queue.
$L_1$ is downloaded first, consolidated, and uploaded.
Since the upload finishes before Step~3, which compacts $L_1$ again and rewrites its frozen part, the result still matches the frozen part on the GPU, so \sys safely swaps it in (green check).
In contrast, the SVD of $L_2$ is still running when Step~3 compacts $L_2$ again, making the result on the CPU stale.
\sys therefore drops this consolidation (red cross) and immediately starts consolidating $L_3$, which is uploaded before its next compaction at Step~4.
After Step~3, $L_2$, $L_4$, and $L_1$ all wait for consolidation.
Since $L_2$ has failed before, \sys retries it first (i.e., dropped-first) and finishes it before its next compaction at Step~5.
By the end of Step~4, $L_4$, $L_1$, and $L_3$ wait for consolidation. 
Since $L_4$ has never been consolidated, \sys prioritizes it (i.e., least-recently-consolidated) and completes its consolidation before Step~6.
In practice, \sys uses a CUDA event to trigger consolidation once the frozen part is ready.
The CPU then manages the data transfers on a separate CUDA stream, performs the SVD, and swaps the pointers, all without stalling decoding on the GPU.
To avoid CPU contention, \sys pins the download, SVD, and upload threads to different CPU cores and sets the number of OpenMP threads for the SVD to the cores available to it.

\subsection{Kernel Design}
\label{sec:design:kernel}

To handle uneven ranks across adaptively compressed states, \sys uses dedicated kernels with a compact memory layout.

\textbf{Memory layout.}
To avoid writing back the state at every decoding step (Section~\ref{sec:design:local_svd_update}), \sys stores the buffer separately from the low-rank factors, so that buffering a new token only appends its key and value to the buffer.
Since an asynchronous copy between the GPU and the CPU operates on a contiguous tensor (Section~\ref{sec:design:async_consolidation}), mixing active and frozen rows would require an extra gather kernel before each transfer, thereby delaying the decoding.
\sys therefore separates the frozen part from the active part and stores each in a contiguous region, so that every transfer can start directly without gathering.
\sys uses a CSR-like format to compactly store the factors of all states, with an offset array that records their starting positions.
After each compaction, \sys reallocates the storage to exactly fit the new ranks and updates the offsets accordingly.
For some states, the low-rank form does not save memory.
For example, a $128\times128$ state of rank $r$ stores $256r$ values in the low-rank form, which exceeds the $128\times128$ values of the dense form when $r>64$.
\sys therefore falls back to the dense form for such states after initialization (Section~\ref{sec:design:local_svd_update}) and stores them in a separate dense region.

\begin{figure*}[htbp]
    \centering
    \includegraphics[width=\linewidth]{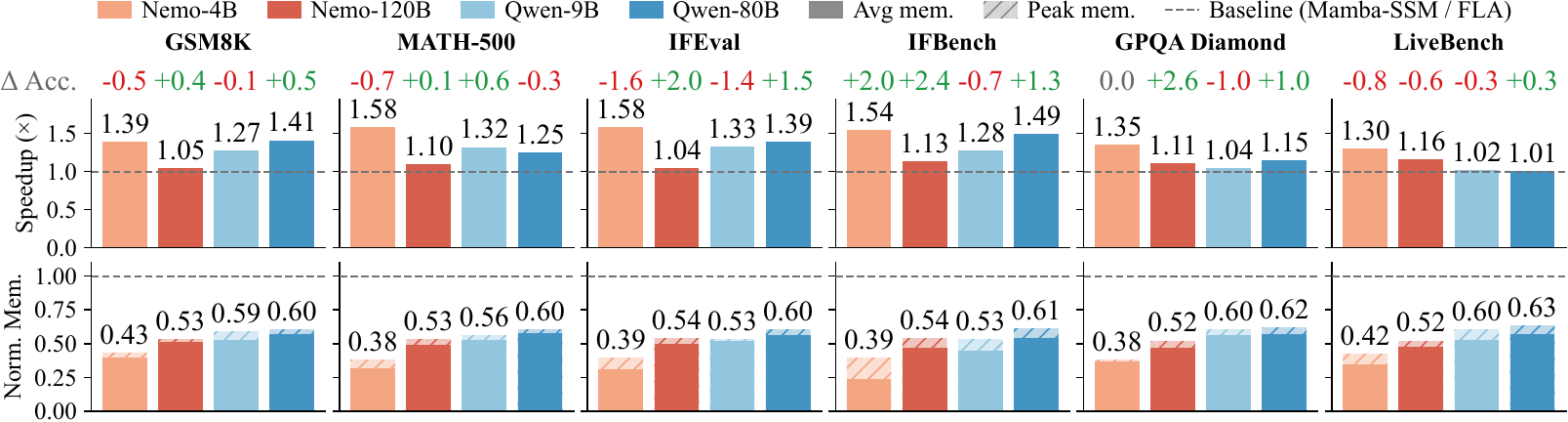}
    \vspace{-16pt}
    \caption{End-to-end performance. Geometric-mean speedup on Nemo-4B,
     Nemo-120B, Qwen-9B, and Qwen-80B: 1.45$\times$, 1.10$\times$,
     1.20$\times$, 1.27$\times$; normalized peak state memory: 0.40, 0.53,
     0.57, 0.61 (i.e., memory saving: 60\%, 47\%, 43\%, 39\%).}
    \label{fig:eval:end_to_end}
\end{figure*}

\textbf{Kernel fusion, load balancing, and support for CUDA graphs.}
Since the ranks are small, running each operation on the factors as a separate kernel would be dominated by launch overhead and intermediate memory traffic.
\sys therefore fuses these operations in two places.
First, a single decoding kernel appends the new key-value pair to the buffer and computes $y_t$ directly from the factors and the buffer, applying the accumulated decay on the fly without materializing the dense state.
Second, \sys fuses chains of algebraic operations on the factors, such as subspace iteration, orthogonalization, and concatenation, thereby avoiding writing intermediate results back to global memory.
Besides, assigning one program to each state leaves programs in low-rank states idle while those in high-rank states are still running.
Therefore, for row-independent operations such as readout, \sys launches one program per component row to balance the workload.
\sys also supports CUDA graphs through a device-side pointer table and a counter.

\section{Evaluation}
\label{sec:eval}

\subsection{Experimental Setup}
\label{sec:eval-setup}

\textbf{Models.}
\sys is evaluated on four hybrid models: Nemotron-3-Nano-4B (Nemo-4B) ~\cite{nvidia2026nemotron3nano4b} and Nemotron-3-Super-120B-A12B (Nemo-120B)~\cite{nvidia2026nemotron3super}, whose recurrent layers are Mamba2~\cite{dao2024mamba2}, and Qwen3.5-9B (Qwen-9B)~\cite{qwen2026qwen35} and Qwen3-Next-80B-A3B-Instruct (Qwen-80B)~\cite{qwen3technicalreport}, who use GDN~\cite{yang2025gdn}.

\textbf{Hardware.}
All experiments run on one NVIDIA RTX PRO 6000 Blackwell GPU with 96~GB of GDDR7 memory, connected via PCIe~5.0 to two Intel Xeon 6515P CPUs (32 cores in total) with 252~GB of host memory.
Nemo-4B and Qwen-9B run in BF16, while Nemo-120B and Qwen-80B use NVFP4 weights with an FP8 KV cache to fit on a single GPU.
The recurrent states are kept in FP32 for all models.

\textbf{Baselines.}
Mamba-SSM~\cite{gu2024mamba} and FLA~\cite{yang2024fla} provide recurrent kernels widely adopted by serving engines such as vLLM~\cite{kwon2023vllm} and SGLang~\cite{zheng2024sglang}.
We therefore use their dense-state implementations as baselines.
Unless otherwise stated, \sys and the baseline use the same batch size for each benchmark, chosen based on available GPU memory.

\textbf{Workloads and metrics.}
End-to-end performance is measured on GSM8K~\cite{cobbe2021trainingverifierssolvemath}, MATH-500~\cite{lightman2024let, hendrycks2021measuring}, IFEval~\cite{zhou2023instruction}, IFBench~\cite{pyatkin2026generalizing}, GPQA Diamond~\cite{rein2023gpqa}, and LiveBench~\cite{white2025livebench} with greedy decoding and the default generation length of each benchmark.
Accuracy on long contexts and long generations is further measured on four NIAH tasks of RULER~\cite{hsieh2024ruler}, and on MATH-500 and GPQA Diamond with thinking enabled and up to 32K generated tokens.
Speedup is the end-to-end serving time of the baseline over that of \sys on each benchmark, and Norm.\ Mem.\ is the memory of the recurrent states under \sys normalized to that under the baseline, at peak and on average.

\textbf{Configuration of \sys.}
\sys uses $\varepsilon=3\times10^{-3}$ for the Nemotron models and $\varepsilon=10^{-2}$ for the Qwen models, a buffer size of $K=8$, and the top eight components as the active subspace for each local SVD.

\subsection{End-to-End Performance}
\label{sec:eval-e2e}

\textbf{\sys speeds up decoding by 1.25$\times$ and nearly halves the
state memory without hurting accuracy.}
Figure~\ref{fig:eval:end_to_end} reports the speedup, the accuracy
change, and the state memory of \sys on the six benchmarks.
\sys speeds up decoding by 1.25$\times$ and by up to 1.58$\times$ (geomean, Nemo-4B on MATH-500 and IFEval).
Meanwhile, the accuracy changes by $-1.6$ to $+2.6$ points with no
consistent direction: it improves in 12 of 24
model-benchmark pairs, and in average $+0.3$ points.
Speedup is the largest on Nemo-4B (1.30--1.58$\times$), whose recurrent
states dominate the step time of the model, and the smallest on
Nemo-120B (1.04--1.16$\times$), 
where the MoE layers take a larger share of each step and the larger weights leave much less memory for large batches.
On the Qwen models, the speedup is lower on GPQA Diamond and LiveBench, whose long outputs keep the batch size small.
At small batches, the state traffic is only a small part of the step
time, so shrinking it brings less benefit (\S\ref{sec:eval-kernel}).
\sys also reduces the peak memory of the recurrent states to 0.38--0.63 of the baseline (i.e., 1.6--2.6$\times$ smaller, 0.52 on average) and the average memory to 0.24--0.58.
The saving is the largest on Nemo-4B (0.38--0.43 at peak) and the
smallest on Qwen-80B (0.60--0.63 at peak), consistent with the higher
ranks of the GDN states (Figure~\ref{fig:frozen_inflation}).

\subsection{Kernel Speedup}
\label{sec:eval-kernel}

\begin{figure}[htbp]
    \centering
    \includegraphics[width=\linewidth]{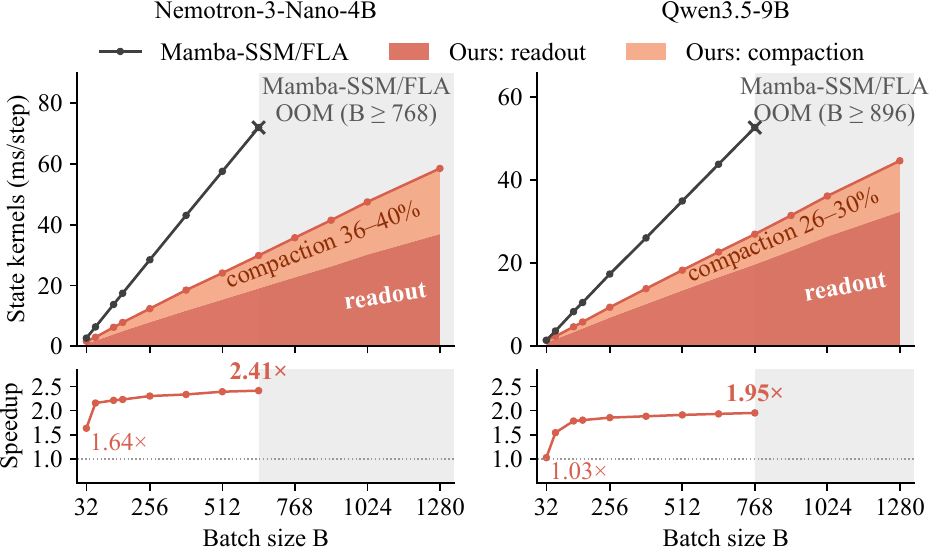}
    \vspace{-16pt}
    \caption{State-kernel time per decoding step (top) and speedup (bottom) under different batch sizes of \sys vs. Mamba-SSM/FLA. Shaded regions: the baselines run out of memory.}
    \label{fig:eval:state_kernel}
\end{figure}

\textbf{\sys cuts the state-kernel time by up to 2.41$\times$ and serves
up to 2$\times$ larger batches.}
Figure~\ref{fig:eval:state_kernel} isolates the time spent in the state
kernels at each decoding step as the batch size grows.
The time of the dense kernels in Mamba-SSM and FLA grows linearly with
the batch size, since every step reads and writes the full dense state
of each request.
\sys grows much more slowly, reducing this time by 1.64--2.41$\times$ on
Nemo-4B and by 1.03--1.95$\times$ on Qwen-9B.
The speedup grows with the batch size and levels off beyond $B=128$:
at small batches, the kernels do not saturate the memory bandwidth, so reducing the traffic helps less.
Of the state-kernel time of \sys, compaction takes 36--40\% on Nemo-4B
and 26--30\% on Qwen-9B, and readout takes the rest.
Since the factors are smaller than the dense states, \sys can serve
larger batches.
The baselines fit at most $B=640$ on Nemo-4B and $B=768$ on Qwen-9B (shaded beyond), while \sys still runs at $B=1280$ on both, i.e., 2$\times$ and 1.67$\times$ larger.
At these larger batches, the state kernels of \sys also take less time
per step than those of the baselines at their largest batch.

\subsection{Accuracy}
\label{sec:eval-accuracy}

\begin{figure}[htbp]
    \centering
    \includegraphics[width=\linewidth]{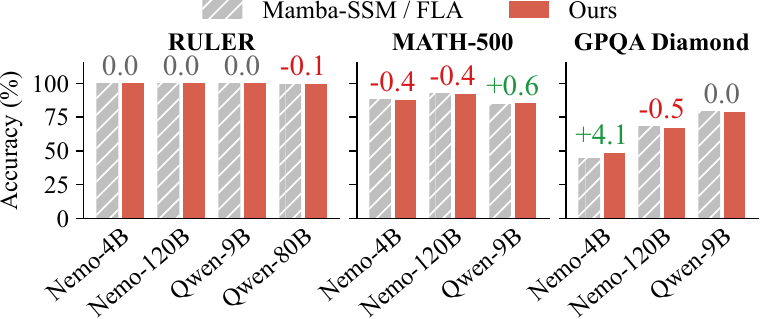}
    \vspace{-16pt}
    \caption{Accuracy on long-context retrieval (RULER) and long reasoning in thinking mode (MATH-500, GPQA Diamond, up to 32K generated tokens). Numbers show the accuracy change of \sys over the baseline.}
    \label{fig:eval:accuracy}
\end{figure}

\textbf{\sys preserves accuracy on long contexts and after 32K
generated tokens.}
Figure~\ref{fig:eval:accuracy} evaluates \sys on long-context
retrieval and on long reasoning with thinking enabled.
On RULER, we report the mean accuracy over four NIAH subsets (single, multi-key, multi-value, and multi-query) with 16K contexts.
\sys matches the baseline on all four models, with at most a 0.1\% drop (Qwen-80B).
On MATH-500 and GPQA Diamond, where an answer takes up to 32K generated
tokens, \sys loses at most 0.5\%.
The 4.1-point gain on GPQA Diamond for Nemo-4B arises because the
baseline reaches the generation limit on more questions.
Qwen-80B is omitted on these two benchmarks because it is an instruct
model.
Thus, local SVD and the frozen tail do not degrade accuracy even after
32K decoding steps.

\subsection{Verbosity}
\label{sec:eval-verbosity}

\begin{figure}[htbp]
    \centering
    \includegraphics[width=\linewidth]{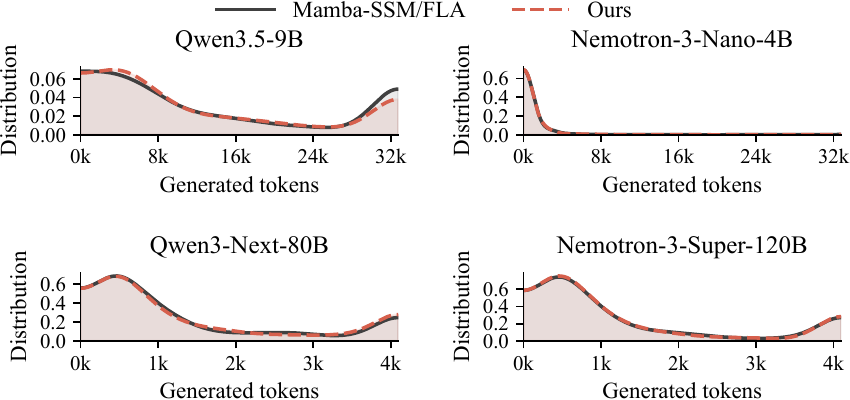}
    \vspace{-16pt}
    \caption{Distribution of the number of generated tokens on MATH-500 under \sys and the baseline.}
    \label{fig:eval:verbosity_math500}
\end{figure}

\textbf{\sys preserves the output length.}
Figure~\ref{fig:eval:verbosity_math500} compares the distributions of
the number of generated tokens on MATH-500.
On Nemo-4B, Nemo-120B, and Qwen-80B, the distributions of \sys and the
baseline nearly overlap.
On Qwen-9B, whose outputs often reach the 32K limit, \sys places
slightly less mass at the limit ($11.4\%$ vs.\ $16.0\%$ of the outputs reach 32K), while the rest of the distribution again overlaps.

\subsection{Ablation Study}
\label{sec:eval-ablation}

\begin{figure}[htbp]
    \centering
    \includegraphics[width=\linewidth]{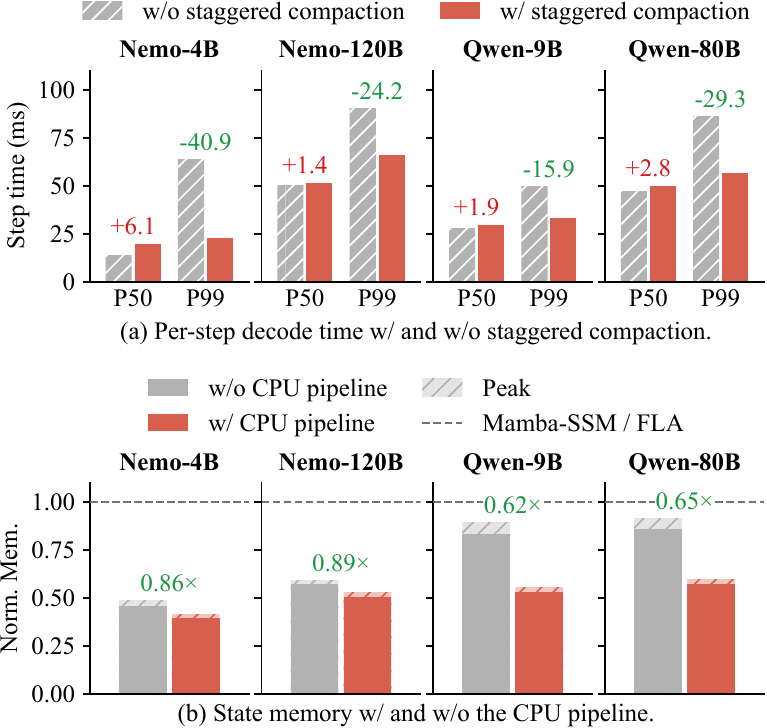}
    \vspace{-16pt}
    \caption{Ablation of (a) staggered compaction and (b) CPU consolidation. Numbers show the step-time change (ms) in (a) and the state-memory ratio of w/ to w/o the CPU pipeline in (b).}
    \label{fig:eval:ablation}
\end{figure}

\textbf{Staggered compaction removes latency spikes.}
Without staggering, all layers compact their buffers at the same step,
causing a latency spike every $K$ steps.
Staggering reduces the P99 step time by 15.9--40.9~ms across the four
models (Figure~\ref{fig:eval:ablation}(a)), e.g., by 2.8 $\times$
on Nemo-4B.
In exchange, the P50 step time increases but only slightly on Nemo-120B,
Qwen-9B, and Qwen-80B (1.4--2.8~ms), since every step now runs a small
part of the compaction.
The increase is larger on Nemo-4B (6.1~ms), whose short step time
makes the compaction share more visible, yet Nemo-4B also gains the
largest P99 reduction (40.9~ms).

\textbf{CPU consolidation is essential for the GDN models.}
Without the CPU pipeline, redundant components keep accumulating in the
frozen part (Figure~\ref{fig:frozen_inflation}).
The CPU pipeline removes them, reducing the state memory by a factor of
0.86 and 0.89 on Nemo-4B and Nemo-120B, and 0.62 and 0.65 on Qwen-9B
and Qwen-80B (Figure~\ref{fig:eval:ablation}(b)).
On the Qwen models, the CPU pipeline contributes more to the memory
saving, since their states show larger rank inflation under local SVD
alone (Figure~\ref{fig:frozen_inflation}).

\subsection{Rank Distribution Visualization}

\begin{figure}[htbp]
    \centering
    \includegraphics[width=\linewidth]{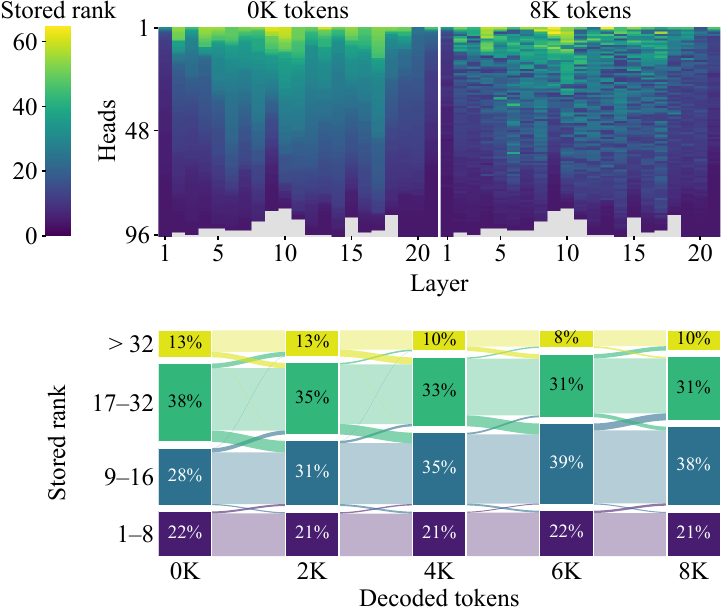}
    \vspace{-16pt}
    \caption{Stored rank per head of Nemo-4B on RULER after prefill and after 8K decoded tokens (top; gray: dense), and the share of heads per rank range (bottom).}
    \label{fig:eval:visualization}
\end{figure}

\textbf{\sys adapts ranks to individual heads.}
Figure~\ref{fig:eval:visualization} shows the ranks stored by \sys for Nemo-4B.
We prefill the first 8K tokens of each of 32 RULER sequences, then teacher-force the remaining 8K one token per step.
Ranks vary widely across heads and layers: 22\% retain at most 8 components, 13\% retain more than 32, and a few use dense states.
A fixed rank would therefore waste memory or lose information.
During decoding, the distribution remains stable: each rank group's share changes by at most 10 percentage points.
Heads shift mainly from 17--32 components ($38\%\to31\%$) to 9--16 components ($28\%\to38\%$).

\section{Related Work}

\paragraph{Inference systems for hybrid models.}
LLM serving systems are built around the token-indexed KV cache, with
continuous batching~\cite{yu2022orca}, IO-aware attention
kernels~\cite{dao2022flashattention}, paged memory
management~\cite{kwon2023vllm}, and prefix
caching~\cite{zheng2024sglang,qin2025mooncake}.
Hybrid serving systems focus on reusing states across
requests.
They store state checkpoints alongside the KV
cache~\cite{huang2026unifiedradix}, decide which checkpoints to keep
and where to place them under a memory
budget~\cite{pan2025marconi,shirokikh2026sparseprefix}, extend
position-independent caching to hybrid
models~\cite{liu2026hypic,liu2026linearkv}, or reconstruct a state by
replaying a recent suffix of its inputs~\cite{liu2026suffixreplay}.
ReplaySSM buffers recent inputs and writes the state back periodically, removing most state writes, but every step reads the full state~\cite{liou2026replayssm}.
Existing systems keep the state dense, whereas \sys shrinks it,
reducing both reads and writes of every step.

\paragraph{Compression for LLM inference.}
Weights are quantized with second-order or activation-aware
methods~\cite{frantar2023gptq,xiao2023smoothquant,lin2024awq} or
factorized with SVD~\cite{yuan2023asvd,wang2025svdllm}, and the KV cache
is compressed by low-bit quantization~\cite{liu2024kivi} or low-rank
projection~\cite{chang2025palu,sun2025shadowkv}.
For SSMs, Quamba quantizes weights and activations~\citep{chiang2025quamba}, and Quamba2 additionally quantizes the cached SSM state~\citep{chiang2025quamba2}.
These tensors are fixed or only appended to, whereas the recurrent
state is rewritten at every step.
LeapQuant quantizes the state once per window but still moves every
entry at every step~\cite{pan2026leapquant}, and SketchSSM approximates
only the reads while keeping the full state for
writes~\cite{kwon2026sketchssm}.
\sys instead keeps the state itself in an adaptive low-rank form
without calibration, shrinking both its memory and its per-step
traffic.

\section{Conclusion}
\label{sec:conclusion}

This paper presents \sys, a serving system that thins the recurrent
states of hybrid models during decoding.
\sys builds on the observation called linear attention sink: the queries, keys, and values
of linear layers concentrate in a stable, low-dimensional subspace, so each recurrent state is effectively low-rank.
\sys therefore keeps each state in an adaptive low-rank SVD form, maintaining it throughout the inference.
\sys then utilizes asynchronous CPU consolidation and fused kernels
to keep this maintenance off the critical path of decoding.
On four hybrid models with Mamba-2 and GDN layers, \sys reduces the
peak memory of recurrent states by up to 2.6$\times$, speeds up
decoding by up to 1.58$\times$, and serves up to 2$\times$ larger
batches, while maintaining model accuracy.

\nocite{langley00}

\bibliography{references}
\bibliographystyle{mlsys2025}

\end{document}